\documentclass{./archive_files/amia}
\usepackage{lipsum} %Remove if not needed
\usepackage{amsmath}
\usepackage{amssymb}
\usepackage{mathtools}
\usepackage{amsthm}
\usepackage{hyperref}
\usepackage{xcolor}
\usepackage{xspace}
\usepackage{balance}
\usepackage{multicol}
\usepackage{multirow}
\usepackage{booktabs}
\usepackage{enumitem}
\usepackage{subfig} 
\usepackage{tcolorbox}
\usepackage[utf8]{inputenc}
\tcbuselibrary{skins}
\usepackage{algorithm}
\usepackage{algorithmic}
\usepackage{caption}
\usepackage{etoolbox}

\newcommand{\header}[1]{\noindent \textbf{\textit{#1}}}

\AtBeginDocument{%
\setlength{\abovedisplayskip}{2pt}
\setlength{\belowdisplayskip}{2pt}
\setlength{\abovedisplayshortskip}{2pt}
\setlength{\belowdisplayshortskip}{2pt}
}

\begin{document}

\title{Enhanced Atrial Fibrillation Prediction in ESUS Patients with Hypergraph-based Pre-training}

% \author{Yuzhang Xie, MS\footnotemark[1], Yuhua Wu, BS\footnotemark[1], Ruiyu Wang, \\ Fadi Nahab, MD, Xiao Hu, PhD, Carl Yang, PhD}
\author{Yuzhang Xie\footnotemark[1], Yuhua Wu\footnotemark[1], Ruiyu Wang, Fadi Nahab, MD, Xiao Hu, PhD, Carl Yang, PhD}
\institutes{
    Emory University, Atlanta, GA, USA; \\
}

\footnotetext[1]{Equal contribution as co-first authors.}

\maketitle

\section*{Abstract}

% \textit{Abstract text goes here, justified and in italics.  The abstract should be in 125-150 words.}

\textit{Atrial fibrillation (AF) is a major complication following embolic stroke of undetermined source (ESUS), elevating the risk of recurrent stroke and mortality. Early identification is clinically important, yet existing tools face limitations in accuracy, scalability, and cost. Machine learning (ML) offers promise but is hindered by small ESUS cohorts and high-dimensional medical features. To address these challenges, we introduce supervised and unsupervised hypergraph-based pre-training strategies to improve AF prediction in ESUS patients. We first pre-train hypergraph-based patient embedding models on a large stroke cohort (7,780 patients) to capture salient features and higher-order interactions. The resulting embeddings are transferred to a smaller ESUS cohort (510 patients), reducing feature dimensionality while preserving clinically meaningful information, enabling effective prediction with lightweight models. Experiments show that both pre-training approaches outperform traditional models trained on raw data, improving accuracy and robustness. This framework offers a scalable and efficient solution for AF risk prediction after stroke.}
\section*{Introduction}

% ESUS
Embolic stroke of undetermined source (ESUS) is an ischemic stroke subtype defined by the absence of an identifiable cause. It is characterized by a non-lacunar brain infarct without an evident embolic source \cite{bhat2021embolic}. 
% AF
Notably, approximately 30\% of ESUS patients develop atrial fibrillation (AF) within six months of their initial stroke, despite having no prior history of AF \cite{elsheikh2024}. This post-ESUS AF is marked by irregular heart rhythms that lead to the formation of embolic clots, thereby increasing the risk of recurrent stroke and hospital readmission \cite{essa2021AF}. Moreover, studies indicate that post-ESUS AF significantly contributes to stroke-related mortality, with death rates surpassing those observed in patients with pre-existing AF \cite{bhatla2021stroke}.
Therefore, accurately predicting AF in ESUS patients is crucial, as it enables clinicians to identify individuals at high risk and implement targeted interventions, ultimately improving the quality of post-stroke care and reducing both mortality and morbidity \cite{kaarisalo1997AF}.

% Traditional Methods - two problems, not comprehensive and efficient
Traditional methods for predicting AF include clinical scoring systems such as C$_2$HEST \cite{li2019AFTraditional}, CHA$_2$DS$_2$-VASc \cite{chen2019AFTraditional}, FHS-AF \cite{schnabel2010AFTradition}, and CHARGE-AF \cite{alonso2013AFTraditional}. 
Although these clinical scoring systems are accessible and easy to implement, they rely on a limited set of population and clinician-selected predictors, which may not capture the full complexity of patient data. 
For instance, the CHARGE-AF score was developed in middle-aged, predominantly Caucasian cohorts \cite{goudis2023AFTraditional}, so it may underestimate AF risk in non-Caucasian patients. The FHS-AF omits ECG and biomarker data, and the CHA$_2$DS$_2$-VASc score relies on a small set of static variables, limiting both tools' ability to capture longitudinal clinical complexity.
These limitations highlight the need for more advanced methods that can efficiently integrate diverse and comprehensive patient data for AF prediction.

% ICD - comprehensive
To address the need for comprehensive patient data in AF prediction, researchers have increasingly utilized International Classification of Diseases (ICD) codes from EHR data. These standardized diagnostic codes offer extensive, structured information about a patient's medical history, comorbidities, and treatment patterns, and thus are valuable for future risk prediction \cite{sarwar2022ICD}.  
ICD codes also enable the identification of meaningful patterns and associations that may not be apparent through traditional clinical scoring systems \cite{dhingra2023ICD}. These advancements are crucial for facilitating more effective AF detection.

% machine learning - efficient, but still struggling with label issues and high dimensionality, which is this study's motivation
Machine learning (ML) has emerged as a powerful tool for analyzing patients' complex clinical data \cite{zhang2022shifting,wang2025biomedjimpact,wu2025utilizing,wu2025towards} and predicting various diseases \cite{xie2024improving,shamout2020machine,xie2025kerap}, including AF. Unlike traditional methods that rely on predefined rules or clinician-selected predictors, ML models excel at processing and analyzing large, complex datasets to uncover patterns beyond human interpretation \cite{bhasuran2025preliminary,xie2024promptlink,xie2022survival}. Researchers have widely applied ML techniques, such as logistic regression, random forests, and gradient boosting trees, to AF prediction tasks \cite{ming2024AF, hart2017AFML, andayeshgar2022AFML}. These methods have achieved impressive performance for AF prediction. However, these traditional ML methods often face significant data issues when applied to medical datasets, including:
\begin{itemize}[topsep=-5pt, itemsep=0pt, parsep=0pt, ]
    \item (1) Sample scarcity: Many medical datasets have limited sample sizes or few labeled examples. This scarcity can make supervised learning impractical, leading to overfitting and reduced generalizability.
    \item (2) High Dimensionality: Comprehensive patient data, such as patients' ICD codes, often involve thousands of features. This ``curse of dimensionality'' increases computational complexity and can cause overfitting, particularly when combined with limited sample sizes.
\end{itemize}

% our key design
The same challenges arise when predicting future AF risks in our cohort of ESUS patients due to the small sample size (510 samples) and high input dimensionality (over 1,000 diagnostic features including ICD). To overcome these challenges, we propose novel methods that leverage pre-training and transfer learning techniques. 
% why hypergraph
We employed hypergraph representation learning because it extends beyond conventional (pairwise) graph methods and shows state-of-the-art performance on EHR-based prediction tasks \cite{xu2023hypergraph,xie2025hypkg}. While a standard graph only models binary relationships between two nodes (e.g., co-occurrence of two ICD codes), a hypergraph allows one hyperedge to connect multiple ICD codes across encounters simultaneously \cite{wu2025abstract,xie2025abstract}. This design captures higher-order and group-level clinical associations that naturally occur in patient trajectories, where diagnoses, medications, and procedures often appear together rather than in isolated pairs \cite{zhang2024tacco}. By leveraging hyperedges, we can better represent the complex, multi-relational structure of patients' ICD codes, improving the ability to learn meaningful patient embeddings for AF prediction.

% Our design
To the best of our knowledge, we are the first to apply hypergraph pre-training for cross-cohort, cross-task transfer (from PSCI to AF prediction) in ESUS. By pre-training on a larger dataset such as AI-RESPECT (7780 samples), with similar features and related labels, we develop generalized models that capture essential patterns. We develop two different hypergraph-based pre-training techniques: (i) supervised pre-training that leverages available labels in the pre-training dataset, and (ii) unsupervised pre-training that uses augmentation-driven objectives to learn robust representations from unlabeled examples. These pre-trained models are then ``transferred'' to our smaller ESUS-AF dataset to generate compact representations for the AF risk prediction task, effectively mitigating data limitations and improving downstream performance.

% Experiments

In our experiments, we compare AF risk prediction using pre-training and transfer learning with traditional ML models applied directly to raw data. Our methods achieve a 5--15\% AUROC improvement and remain robust in ablation studies and external validation. These findings highlight the scalability of hypergraph-based pre-training for complex clinical datasets facing sample scarcity and high dimensionality. In practice, this framework enables more efficient and accurate identification of high-risk ESUS patients for timely intervention and closer follow-up.
\section*{Data Description}
\label{sec:data}
This study leverages comprehensive EHRs from three datasets.
\begin{itemize}[topsep=-5pt, itemsep=0pt, parsep=0pt, partopsep=-3pt, ]
    \item Target dataset: The ESUS-AF dataset contains patients with ESUS and serves as the target cohort with labeled AF outcomes.
    \item Pre-training dataset: The AI-RESPECT dataset is a larger stroke cohort labeled for post-stroke cognitive impairment (PSCI) and used for pre-training and transfer learning.
    \item External validation dataset: The MIMIC-IV dataset provides a large, diverse set of stroke patients labeled with ESUS and AF for external validation of AF prediction.
\end{itemize}

The ESUS-AF dataset includes 510 patients diagnosed between January 1, 2015, and December 13, 2023, of whom 107 (21.0\%) developed post-stroke AF within the timeline. Inclusion criteria require patients to be at least 18 years old, have no documented stroke within a five-year retrospective window prior to 2015, and have no history of AF before the index stroke. For the ESUS-AF dataset, we extract 53 baseline categorical or numerical features from the literature as candidate predictors, including (I) 6 demographic variables (II) 9 blood biomarkers, (III) 21 electrocardiographic (ECG) features, and (IV) 17 comorbidities of stroke.
In addition, all prior-stroke diagnoses were collected, yielding 1,529 binary diagnostic features, including 990 ICD-based features and 539 features about prescribed medications.

To enhance performance, we pre-train on the larger AI-RESPECT cohort (7,780 stroke patients, 1,735 PSCI cases, 2012–2021) and transfer to ESUS-AF. AI-RESPECT requires a stroke diagnosis without prior cognitive impairment and provides 2,595 diagnostic features. Across the AI-RESPECT cohort and ESUS-AF cohort, 1,494 diagnostic features overlap, covering 97.71\% of the ESUS dataset’s diagnostic features.

For external validation, we use MIMIC-IV \cite{johnson2023mimic}, identifying 1,717 potential ESUS patients via ICD-10 code I63.4 (cerebral infarction due to unspecified occlusion or stenosis of cerebral arteries), which is commonly used as a proxy for ischemic stroke of undetermined source. Among these patients, 248 have binary prior-stroke AF (ICD-10 I48 atrial fibrillation and flutter). Each patient consists of 34 baseline features and 1,057 stroke-prior diagnostic features derived from longitudinal medical records, capturing comorbidities and prior history.
\section*{Methods}
\label{sec:method}

\header{Problem Formulation.}
Let $\mathcal{D}_{\text{tg}} = \{(x^{(i)}, y^{(i)})\}_{i=1}^{n}$ represent our target dataset (ESUS-AF) with $n=510$ patients, where each patient $i$ is characterized by a feature vector $x^{(i)} \in \mathbb{R}^{d}$ and a binary AF label $y^{(i)} \in \{0, 1\}$ indicating absence or presence of AF. Since $n$ is relatively small, we leverage transfer learning to enhance feature representations. Specifically, $x^{(i)}$ is constructed using methods such as transfer learning, integrating information from two sub-vectors: (1) a baseline clinical feature vector $x^{(i)}_{b} \in \mathbb{R}^{d_b}$ and (2) a diagnostic feature vector $x^{(i)}_{di} \in \mathbb{R}^{d_{di}}$, where dimension $d_b = 53$ and $d_{di} = 1529$. The objective of this binary classification task is to generate prediction $\hat{y}^{(i)}\in \{0,1\}$.

\begin{figure*}[htbp]
    \centering
    \includegraphics[width=0.9\linewidth]{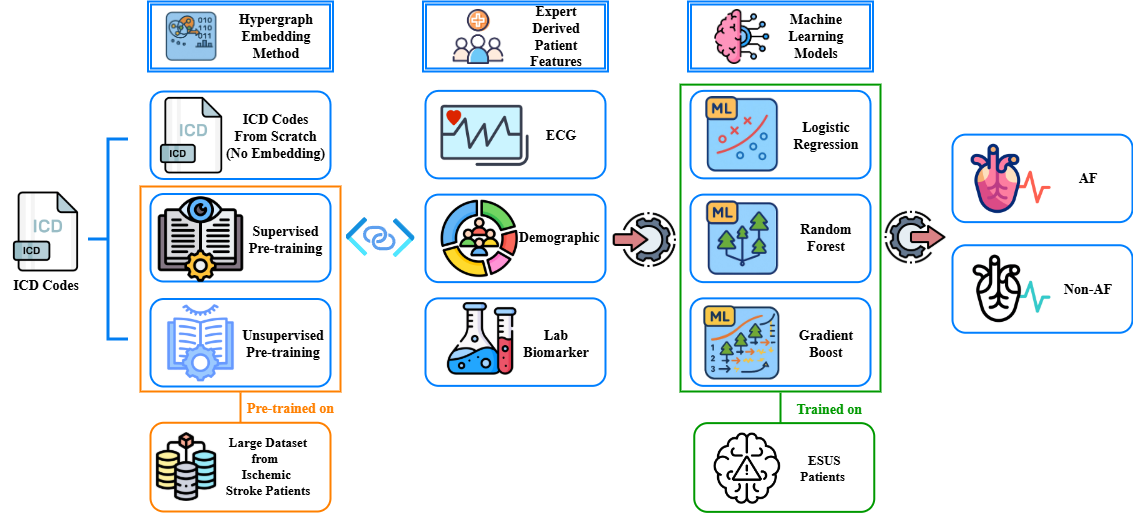}
    \captionsetup{font=footnotesize} 
    \caption{Overview of our methodology. ICD codes from ESUS patients are used to generate patient representations through three approaches: From Scratch (Empty Embedding), Supervised Pre-training, and Unsupervised Pre-training using a large ischemic stroke cohort. These embeddings are combined with expert-derived features (ECG, demographics, and lab biomarkers) and used to train machine learning classifiers (logistic regression, random forest, and gradient boosting) to predict post-stroke atrial fibrillation (AF) versus non-AF outcomes.}
    \label{fig:framework}
\end{figure*}

\header{Overview.}
The overall methodology framework is shown in Figure \ref{fig:framework}. Given the sample scarcity (\(n=510\)) and high dimensionality (53 baseline features and 1,529 diagnostic features) of our ESUS-AF patient dataset, we propose novel methods leveraging transfer learning and pre-training techniques. To evaluate different learning strategies for AF prediction, we obtain patient feature representations using three approaches: (I) From-scratch, which directly uses the raw features; (II) Supervised, which pre-trains a hypergraph model on a larger labeled dataset and then transfers the pre-trained model to learn features to our target dataset; and (III) Unsupervised, which pre-trains a hypergraph model without using labels. Each representation is then fed into three AF prediction classifiers: Logistic Regression (LR), Random Forest (RF), and Gradient Boosting (GB).

\header{From Scratch.}
We construct each patient representation $x^{(i)}$ for patient $i$ by concatenating the baseline feature vector $x^{(i)}_{b}$ with the diagnostic feature vector $x^{(i)}_{di}$:
\begin{equation}
\label{eq:feature_combination}
    x^{(i)} = x^{(i)}_{b} \oplus x^{(i)}_{di}.
\end{equation}

A predictive model $M_{\text{AF}}$ (LR/RF/GB) is trained on these concatenated features to produce a predicted label for AF:
\begin{equation}
\label{eq:af_prediction}
    \hat{y}^{(i)} = M_{\text{AF}}\bigl(x^{(i)}\bigr).
\end{equation}
% The algorithm of this method is described in Appendix \ref{app:from_scratch}. 
This method is purely data-driven within the ESUS-AF cohort and does not use pre-trained models or external datasets. Although straightforward to implement, it is susceptible to overfitting in high-dimensional, small-sample settings and cannot leverage external knowledge.

\header{Supervised Hypergraph-based Pre-training.}
Our proposed supervised method, leveraging hypergraph-based pre-training and transfer learning, involves pre-training a model on larger external datasets (AI-RESPECT), which contain labeled samples to capture generalizable patterns before fine-tuning on the ESUS cohort. Specifically, the pre-training dataset is defined as $\mathcal{D}_{\text{pt}} = \{(x^{(p)}, y^{(p)})\}_{p=1}^{n_p}$. For the pre-training dataset AI-RESPECT, $x^{(p)}$ consists of 2,595 diagnostic features, $n_p = 7,780$ represents the number of samples, and $y^{(p)}$ indicates whether a patient develops PSCI. The overall methodology leverages the structural similarities between the pre-training and target datasets—particularly the high overlap in diagnostic features—by pre-training a model $M_{SP}$ on $\mathcal{D}_{\text{pt}}$. For pre-training, we adopt the state-of-the-art hypergraph transformer method inspired by Xu et al. \cite{xu2023hypergraph}, which formulates EHR-based diagnosis prediction as a hypergraph-based learning problem.

\begin{figure*}[htbp!]
    \centering
    \captionsetup{font=footnotesize} 
    \includegraphics[width=0.7\linewidth]{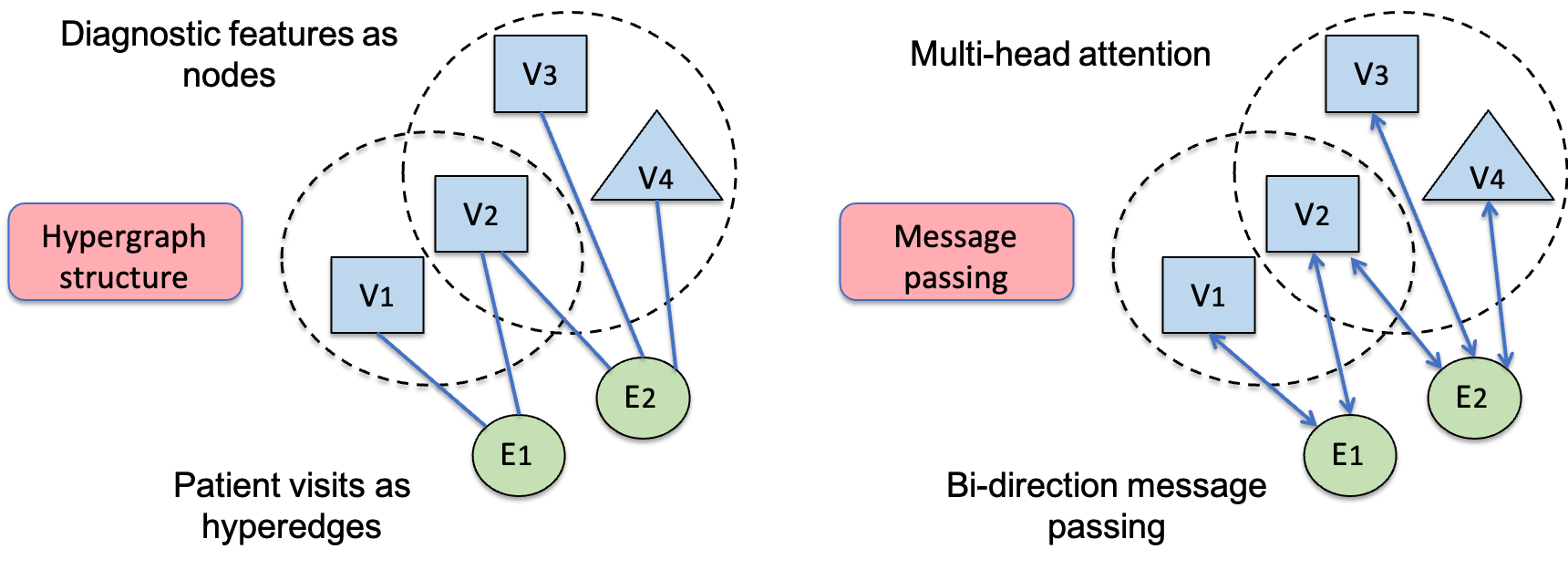}
    \caption{Hypergraph structure and message passing. The hypergraph represents each patient visit as a hyperedge linking multiple diagnostic features, while features serve as nodes shared across visits. During bi-directional message passing, nodes aggregate information from their connected hyperedges, and hyperedges update their embeddings from incident nodes. This allows the model to capture higher-order relationships among diagnostic features and visits.}
    \label{fig:hypergraph}
\end{figure*}

We represent the diagnostic data as a hypergraph $HG = (V, E)$, as illustrated in Figure \ref{fig:hypergraph}, where vertices correspond to diagnostic features and hyperedges represent patient visits. Existing diagnostic features from a patient’s past history (recorded as 1) are included in that patient’s hyperedge. A hypergraph transformer model $M_{SP}$ is then pre-trained using multi-head self-attention to propagate information between connected nodes and hyperedges, following a message-passing framework. At layer $l$, the message propagation (for the hyperedge corresponding to patient $p$) is defined as 
\begin{equation} \label{eq:hg_propogation} x_{e}^{l} = f_{V \to E}(x_{v}^{\,l-1}), \quad x_{v}^{\,l} = f_{E \to V}(x_{e}^{\,l}), 
\end{equation} 
where $x_{e}^{l}$ represents the hyperedge embeddings in the hypergraph at layer $l$, $x_{v}^l$ represents the node embeddings in the hypergraph, $f_{V \to E}$ represents the self-attention mechanism within a hyperedge's connected nodes, and $f_{E \to V}$ represents the self-attention mechanism within a node's connected hyperedges. To implement the message-passing functions $f_{V\to E}$ and $f_{E\to V}$, we employ multi-head self-attention. We use $S$ to denote a set, which may be either a hyperedge (the set of its incident nodes) or a node (the set of its incident hyperedges). If $S$ is a hyperedge, its embedding is updated from its connected node embeddings of the previous layer; if $S$ is a node, its embedding is updated from its connected hyperedge embeddings of the same layer. Let $x_S \in \mathbb{R}^{|S|\times d_{hi}}$ be the input embedding matrix for the set $S$; the updated matrix $x'_S$ is computed as
\begin{equation}
\label{eq:multihead}
x'_S = \text{Concat}_{h=1}^{H} \mathrm{Attention}_h(x_S), \quad \mathrm{Attention}_h(x_S) = \mathrm{softmax}\!\left(\frac{Q_h K_h^\top}{\sqrt{d_k}}\right) V_h,
\end{equation}
and for head $h$,
\begin{equation}
Q_h = x_S W_{Q,h},\quad K_h = x_S W_{K,h},\quad V_h = x_S W_{V,h},
\end{equation}
here $W_{Q,h},W_{K,h},W_{V,h}\in\mathbb{R}^{d_{hi}\times d_k}$ are learnable projection matrices, $d_k=\lfloor d_{hi}/H\rfloor$ is the per-head dimension, and $H$ is the number of heads. Each head produces an output of $|S|\times d_k$; concatenating the $H$ heads yields a matrix of $|S|\times d_{hi}$, which is typically followed by a linear projection, a residual connection, and layer normalization.

During pre-training on $\mathcal{D}_{\mathrm{pt}}=\{(x^{(p)}, y^{(p)})\}_{p=1}^{n_p}$ we optimize the standard cross-entropy loss for the labeled prediction task:
\begin{equation}
\label{eq:app:ce}
\mathcal{L}_{\mathrm{pt}} = -\frac{1}{n_p}\sum_{p=1}^{n_p} \sum_{c} y^{(p)}_{c}\log \hat{y}^{(p)}_{c},
\end{equation}
where $\hat{y}^{(p)} = \mathrm{softmax}(g(x^{(p)}_{e}))$ is the prediction logit from the final hyperedge embedding $x^{(p)}_{e}\!=\!x_{e}^{(p,L)}$. The hypergraph transformer model $M_{SP}$ is optimized on $\mathcal{D}_{\text{pt}}$ and is transferred to $\mathcal{D}_{\text{tg}}$ for generating pre-trained patient representation $x^{(i)}_{tr}$ for patient $i\in \mathcal{D}_{\text{tg}}$ via
\begin{equation}
\label{eq:transfer0}
x^{(i)}_{tr} = x_{e}^{(i,L)} = M_{SP}(x^{(i)}_{di}),
\end{equation}
where $M_{SP}$ denotes the pre-trained model, $x^{(i)}_{di}$ denotes the diagnostic features for patient $i$, and $x_{e}^{(i,L)}$ denotes the final hyperedge embedding corresponding to patient $i$ in the final layer $L$ after training. The final patient representation $x^{(i)}$ is constructed by concatenating our pre-trained 32-D embedding $x^{(i)}_{tr}$ with the baseline clinical feature vector $x^{(i)}_{b}$:
\begin{equation}
\label{eq:transfer1}
x^{(i)} = x^{(i)}_{b} \oplus x^{(i)}_{tr}.
\end{equation}

The combined representation $x^{(i)}$ is subsequently used as input to the machine learning model $M_{AF}$ for atrial fibrillation (AF) prediction, following the formulation in Eq. (\ref{eq:af_prediction}). This approach aims to enhance predictive performance and generalizability, particularly within the small-sample ESUS-AF cohort.

\header{Unsupervised Hypergraph-based Pre-training.}
Our proposed unsupervised method aims to learn robust patient representations from unlabeled diagnostic data through self-supervised pre-training, before transferring the learned knowledge to the ESUS-AF cohort. Specifically, we pre-train on an external dataset $\mathcal{D}_{\text{pt}}$ (AI-RESPECT) and use only the diagnostic features (without outcome labels). Then we transfer the pre-trained model to the downstream target dataset $\mathcal{D}_{\text{tg}}$. 
Similar to the supervised setting, we represent the data as a hypergraph $HG = (V, E)$, where vertices $V$ denote diagnostic features and hyperedges $E$ correspond to patient visits. The unsupervised model $M_{UP}$ adopts the same architecture and message-passing mechanism as its supervised counterpart; the key distinction lies in the learning objective. While the supervised model learns clinically predictive representations by training on labeled EHR prediction tasks, the unsupervised approach instead constructs two augmented hypergraph views and optimizes the node representations through a self-supervised contrastive objective inspired by prior work~\cite{song2023genSim,lee2022m}. This enables $M_{UP}$ to acquire clinically meaningful patient features without relying on outcome labels.

When generating two different hypergraph views $HG(A)$ and $HG(B)$, each view is stochastically perturbed by applying node masking and hyperedge selection from the original hypergraph $HG$. For node masking, nodes with high duplication scores—quantified based on their connectivity patterns—are more likely to be masked. We define the duplication of a vertex $v$ and its corresponding masking probability jointly as
\begin{equation}
D(v) = \frac{\sum_{e \in E(v)} |e|}{\big|\bigcup_{e \in E(v)} e \big|},
\quad
p_{v} =
\min \left\{
\frac{w_{\max} - w_v}{w_{\max} - w_{\text{avg}}} \cdot p_{\text{node}}, \ 
p_{\tau}
\right\},
\end{equation}
where $E(v)$ is the set of hyperedges containing $v$, $|e|$ denotes the number of vertices in hyperedge $e$, and the denominator of $D(v)$ counts the unique vertices that co-occur with $v$ across its hyperedges. A larger $D(v)$ indicates that $v$ appears in many overlapping visits and is therefore less informative. The masking weight is given by $w_v = \log D(v)$, with $w_{\max}$ and $w_{\text{avg}}$ denoting the maximum and average log-duplication across all vertices, $p_{\text{node}}$ controlling the overall masking probability, and $p_{\tau}$ capping the maximum masking rate.
For hyperedge selection, each hyperedge $e \in E$ is stochastically retained, removed, or modified using Gumbel\mbox{-}Softmax sampling~\cite{jang2016categorical}. 
Given the final embedding $x_e$ obtained from the hypergraph transformer (aggregating node and hyperedge features), we first compute the logits for the three augmentation operations---\emph{preserve}, \emph{remove}, and \emph{mask nodes inside}---and then apply Gumbel\mbox{-}Softmax to sample the operation:
\begin{equation}
\alpha_e = W x_e \in \mathbb{R}^3,
\quad
g_{e,k} = 
\frac{\exp\!\left((\alpha_{e,k} + \gamma_{e,k})/\tau_g\right)}
{\sum_{j=1}^{3} \exp\!\left((\alpha_{e,j} + \gamma_{e,j})/\tau_g\right)},
\quad k \in \{1,2,3\},
\end{equation}
where $\gamma_{e,k} \sim \mathrm{Gumbel}(0,1)$ are i.i.d.\ noise samples and $\tau_g$ is a temperature controlling the sharpness of the distribution. The resulting vectors $\{g_e\}_{e \in E}$ specify the augmentation operations for the perturbed views. 

After generating two augmented views $HG(A)$ and $HG(B)$ of the original hypergraph, we impose five self-supervision objectives to encourage the augmented views to preserve structural information. First, a node-level consistency loss, $\mathcal{L}_{\text{sim}}$, enforces alignment between the embeddings of each node across the two views:
\begin{equation}
\mathcal{L}_{\text{sim}} = \operatorname{MSE}\big(x^{(A)}_v, x^{(B)}_v\big),
\end{equation}
where $x^{(A)}_v$ and $x^{(B)}_v$ denote the embeddings of the same vertex $v$ in views $A$ and $B$, respectively.

Complementing this local constraint, a hyperedge-level contrastive objective, $\mathcal{L}_{\text{hyper}}$, preserves the global topological structure by encouraging consistency between the hyperedge representations of the two views:
% \begin{equation}
% \mathcal{L}_{\text{hyper}} 
% = \tfrac{1}{2}\Big( \ell\big(X^{(A)}, X^{(B)}\big) + \ell\big(X^{(B)}, X^{(A)}\big) \Big),
% \end{equation}
\begin{equation}
\label{eq:hyper_contrastive}
L_{\mathrm{hyper}} = \ell\!\left(X^{(A)}, X^{(B)}\right)
= \frac{1}{|E|}\sum_{e\in E} \ell\!\left(x_e^{(A)}, x_e^{(B)}\right),
\end{equation}
where $X^{(A)}$ and $X^{(B)}$ are the hyperedge embedding matrices for the two augmented views. 
The function $\ell$ denotes a contrastive loss with temperature parameter $\tau_{\text{level}}$:
\begin{equation}
\ell\big(x^{(i)}, x^{(j)}\big)
= - \log
\frac{
\exp\!\left( \operatorname{sim}\!\left(x^{(i)}, x^{(j)}\right) / \tau_{\text{level}} \right)
}{
\sum_{q}
\exp\!\left( \operatorname{sim}\!\left(x^{(i)}, x^{(q)}\right) / \tau_{\text{level}} \right)
},
\end{equation}
where $q$ ranges over all nodes (or hyperedges, depending on the context) in the mini-batch, and $\operatorname{sim}(\cdot,\cdot)$ is a similarity function (e.g., cosine similarity).

In addition, we define contrastive objectives at the node, hyperedge, and membership levels. 
For the node-level contrast, embeddings of the same node across views form positive pairs, while embeddings of other nodes in the mini-batch are treated as negatives. 
Given $x^{(A)}_v$ from $HG(A)$ and its positive counterpart $x^{(B)}_v$ from $HG(B)$, the node-level contrastive loss is
\begin{equation}
L_n = \frac{1}{2|V|}\sum_{v=1}^{|V|} \Big[ \ell\big(x^{(A)}_v, x^{(B)}_v\big) + \ell\big(x^{(B)}_v, x^{(A)}_v\big) \Big].
\end{equation}

Similarly, for each hyperedge $e$, the embeddings $x^{(A)}_e$ and $x^{(B)}_e$ form a positive pair, while $x^{(B)}_{e'}$ with $e' \neq e$ are treated as negatives:
\begin{equation}
L_e = \frac{1}{2|E|} \sum_{e=1}^{|E|} \Big[ \ell\big(x^{(A)}_e, x^{(B)}_e\big) + \ell\big(x^{(B)}_e, x^{(A)}_e\big) \Big].
\end{equation}

% Let $A_{ve} \in \{0,1\}$ denote the incidence matrix indicating whether node $v$ belongs to hyperedge $e$. 
Let $A\in\{0,1\}^{|V|\times|E|}$ be the incidence matrix of the hypergraph, where
$A_{ve}=1$ iff node $v\in V$ belongs to hyperedge $e\in E$, and $A_{ve}=0$ otherwise.
The membership-level contrast aligns associated node–hyperedge pairs across views:
\begin{equation}
L_m = \frac{1}{2K} \sum_{v=1}^{|V|} \sum_{e=1}^{|E|} A_{ve} \Big[ \ell\big(x^{(A)}_v, x^{(B)}_e\big) + \ell\big(x^{(B)}_v, x^{(A)}_e\big) \Big],
\end{equation}
where $K$ is the total number of positive memberships, i.e., the number of nonzero entries in $A$.

Intuitively, these objectives jointly encourage (i) the same diagnostic features to have consistent embeddings across augmented views, (ii) visits with similar diagnostic profiles to be close in the embedding space, and (iii) node–hyperedge incidence patterns to be preserved. The overall pre-training loss is:
\begin{equation}
L_{\text{total}} = L_{\text{sim}} + L_{\text{hyper}} + L_n + L_e + L_m.
\end{equation}
Similar to the supervised pre-training method, after pre-training on $\mathcal{D}_{\text{pt}}$, we transfer the model $M_{UP}$ to the ESUS-AF dataset $\mathcal{D}_{\text{tg}}$ to generate patient-level embeddings $x^{(i)}_{tr}$ for each patient $i$, similar to Eq. (\ref{eq:transfer0}):
\begin{equation}
\label{eq:transfer2}
x^{(i)}_{tr} = x_{e}^{(i,L)} = M_{UP}(x^{(i)}_{di}),
\end{equation}
where $x^{(i)}_{di}$ denotes the diagnostic features of patient $i$ and $x_{e}^{(i),L}$ denotes the final-layer hyperedge embedding. We then concatenate this pre-trained 32-D representation with the baseline clinical feature vector $x^{(i)}_b$, similar to Eq. (\ref{eq:transfer1}):
\begin{equation}
\label{eq:transfer3}
x^{(i)} = x^{(i)}_b  \oplus x^{(i)}_{tr}.
\end{equation}

This hybrid embedding $x^{(i)}$ is subsequently used for downstream AF prediction, enabling the model to leverage rich structural semantics learned from unlabeled patient data.

\section*{Results}
\header{Implementation.}
In the preprocessing stage, missing numerical data within baseline features $x_b$ are imputed using a SimpleImputer with median, and the numerical variables are normalized with MinMaxScaler. We compare three methods for generating the representations for the target ESUS-AF patient: From-scratch, Supervised Hypergraph Pre-training, and Unsupervised Hypergraph Pre-training. We evaluate the performance of these representations by training downstream AF prediction models $M_{AF}$, including logistic regression (LR), random forest (RF), and gradient boosting (GB). When building the AF prediction model $M_{AF}$, the dataset is split into 80\% for training-validation and 20\% for testing. 
Finally, a five-fold nested cross-validation strategy is employed for robust evaluation, with performance measured by AUROC (the primary metric), accuracy, F1-score, and PR-AUC.
The implementation details and code for our experiments are provided in a repository for review on Github \footnote{Github: \href{https://github.com/JonathanWry/Enhanced-Atrial-Fibrillation-Prediction-in-ESUS-Patients-with-Pre-training-and-Transfer-Learning}{https://github.com/JonathanWry/Enhanced-Atrial-Fibrillation-Prediction-in-ESUS-Patients-with-Pre-training-and-Transfer-Learning}}.

\header{AF Prediction Performance.}

We first leverage the AI-RESPECT dataset as a pre-training dataset and then transfer the pre-trained model $M_{SP}$ to the ESUS-AF dataset for the AF prediction task. The five-fold cross-validation performance of the compared methods is summarized in Table \ref{tab:model_comparison}. 

\vspace{10pt}
\begin{table*}[htbp!]
    \centering
    \captionsetup{font=footnotesize} 
    \caption{AF prediction performance comparison for different methods using the ESUS-AF dataset.}
    \label{tab:model_comparison}
    \resizebox{0.7\linewidth}{!}{%
    \begin{tabular}{c c c c c c}
        \toprule
        \textbf{Embedding Method} & \textbf{ML Model} & \textbf{AUROC} & \textbf{Accuracy} & \textbf{F1-score} & \textbf{PR-AUC} \\
        \midrule
        \multirow{3}{*}{From Scratch}
            & LR & 0.489$\pm$0.025 & 0.690$\pm$0.029 & 0.199$\pm$0.077 & 0.254$\pm$0.043 \\
            & RF & 0.494$\pm$0.081 & 0.441$\pm$0.214 & 0.210$\pm$0.152 & 0.232$\pm$0.027 \\
            & GB & 0.512$\pm$0.033 & 0.729$\pm$0.014 & 0.169$\pm$0.071 & 0.246$\pm$0.020 \\
        \midrule
        \multirow{3}{*}{Supervised}
            & LR & 0.617$\pm$0.033 & 0.721$\pm$0.034 & \textbf{0.407$\pm$0.037} & 0.319$\pm$0.046 \\
            & RF & \textbf{0.625$\pm$0.045} & \textbf{0.784$\pm$0.022} & 0.384$\pm$0.111 & \textbf{0.374$\pm$0.042} \\
            & GB & 0.583$\pm$0.036 & 0.759$\pm$0.048 & 0.248$\pm$0.068 & 0.284$\pm$0.010 \\
        \midrule
        \multirow{3}{*}{Unsupervised}
            & LR & 0.616$\pm$0.025 & 0.625$\pm$0.023 & 0.356$\pm$0.036 & 0.312$\pm$0.038 \\
            & RF & 0.620$\pm$0.041 & 0.747$\pm$0.042 & 0.331$\pm$0.069 & 0.321$\pm$0.036 \\
            & GB & 0.582$\pm$0.020 & 0.741$\pm$0.049 & 0.319$\pm$0.062 & 0.292$\pm$0.036 \\
        \bottomrule
    \end{tabular}%
    }
\\
\begin{flushleft}
\footnotesize{Note: ``Embedding Method'' describes how patient representations are generated. ``ML Model'' specifies the AF prediction model (``LR'': logistic regression, ``RF'': random forest, ``GB'': gradient boosting tree). Values are mean$\pm$standard deviation over 5 runs. Bold indicates the best value within each embedding group.}
\end{flushleft}
\end{table*}

\vspace{-10pt}
Table~\ref{tab:model_comparison} demonstrates that both supervised and unsupervised pre-training consistently improve AF prediction performance compared with training from scratch. Supervised pre-training yields performance gains across all metrics and all models (e.g., AUROC improves by 7–12\%, and F1-score improves by 7–20\%). Similarly, unsupervised pre-training also provides substantial improvements. These gains arise because the ESUS dataset is relatively small and high-dimensional, making models trained from scratch more susceptible to overfitting. In contrast, both forms of pre-training introduce informative structure into the embedding space, capture meaningful patient-feature interactions, and produce smoother, more stable representations that lightweight downstream models can better exploit.

Furthermore, supervised pre-training offers stronger task-specific guidance by aligning the embedding space with AF-related labels in a large cohort, allowing the model to learn disease-relevant patterns such as cardiovascular risk factors and comorbidity signatures. Unsupervised pre-training, while lacking such task alignment, captures general clinical structure and remains advantageous in settings without labels. As a result, supervised pre-training typically produces more discriminative representations and higher AUROC, F1-score, and PR-AUC, while unsupervised approaches remain more broadly applicable when labeled data are limited or unavailable.

\header{Ablation Studies.}
\begin{figure*}[htbp!]
    \centering
    \captionsetup{font=footnotesize} 
    \includegraphics[width=0.6\linewidth]{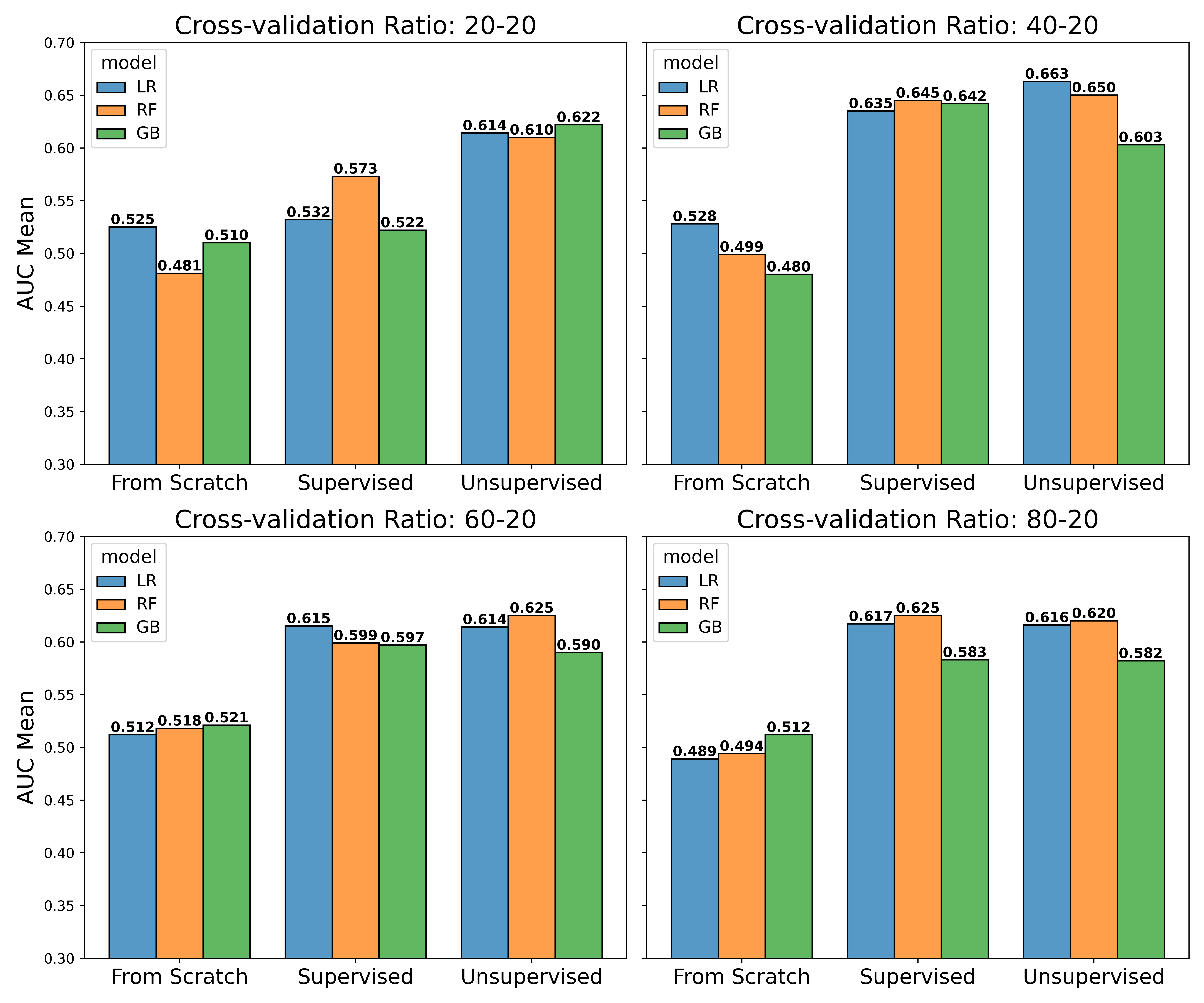}
    \caption{AF prediction performance comparison of different methods under varying training data sizes. The figure reports AUROC across four cross-validation settings (20–20, 40–20, 60–20, and 80–20), where the first number denotes the proportion of data used for training and the second denotes the fixed 20\% test set.}
    \label{fig:ablation_size}
\end{figure*}

To further investigate how the amount of available training data influences the effectiveness of our pre-training strategies, we train the downstream models using different proportions of the training set (20\%, 40\%, 60\%, and 80\%) while keeping the test set fixed to 20\% of the full dataset. We then measure the AUROC under each setting. As illustrated in Figure~\ref{fig:ablation_size}, both supervised and unsupervised pre-training exhibit remarkable robustness: their performance remains stable even when the downstream models are trained with substantially limited data. Across all training ratios, the pre-trained representations consistently outperform models trained from scratch, demonstrating the strong data efficiency of the learned embeddings. Moreover, the performance gap between pre-trained models and the from-scratch baseline becomes larger as additional training data are provided. This trend suggests that larger downstream training sets allow the transferred representations to be further refined, enabling the model to better exploit the structural knowledge encoded during pre-training. In other words, pre-training offers a strong foundation that provides immediate benefits in low-data regimes, while also scaling effectively and delivering even greater gains as more data become available.

\begin{figure*}[htbp!]
    \centering
    \captionsetup{font=footnotesize} 
    \includegraphics[width=0.45\linewidth]{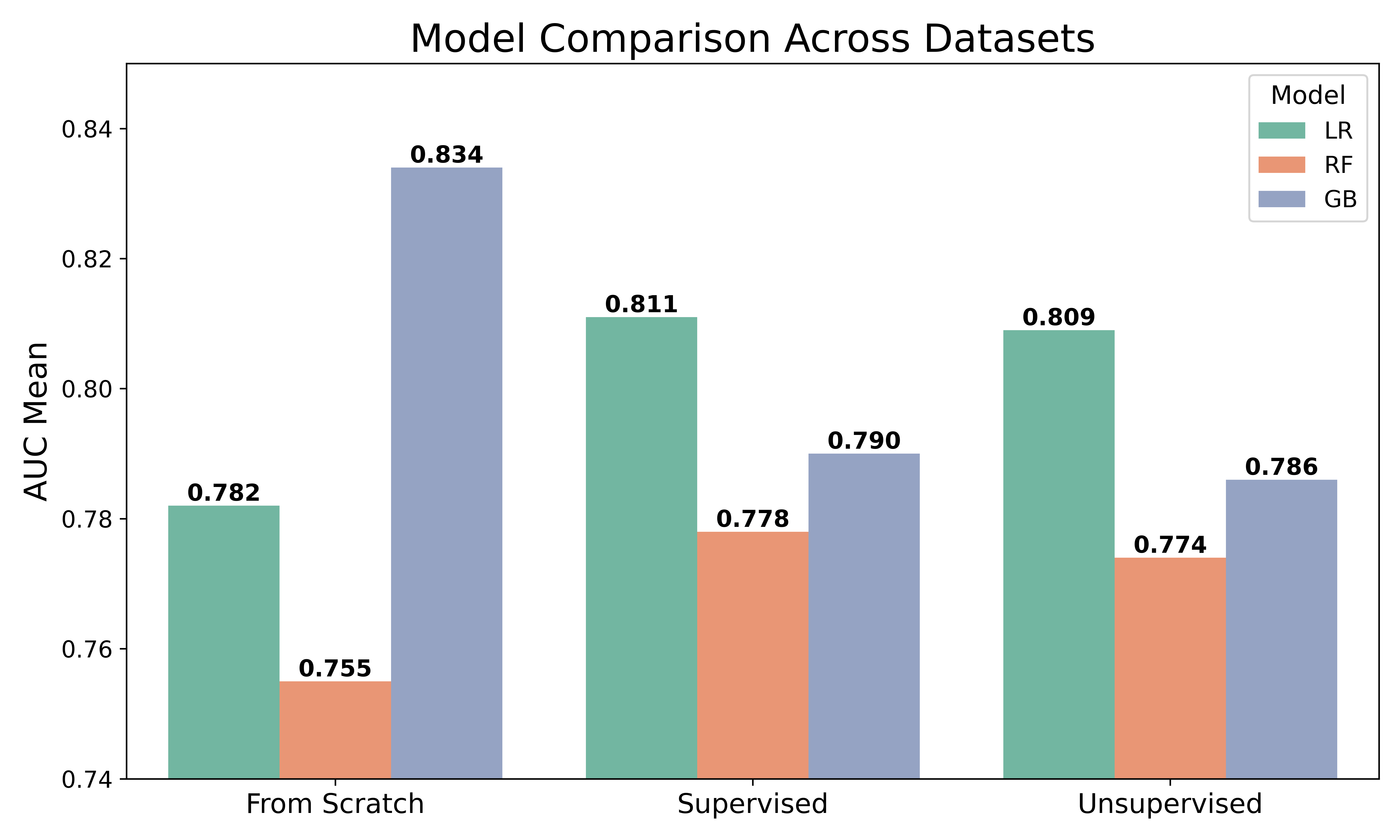}
    \caption{AF prediction performance comparison of different methods on the external validation dataset.}
    \label{fig:mimic}
\end{figure*}

We further evaluate the generalizability of our pre-training method by applying it to an external cohort from the MIMIC-IV dataset. Specifically, we use the model pre-trained on the AI-RESPECT dataset to generate patient-level representations for MIMIC-IV patients, and then train AF prediction models for comparison. This cross-dataset transfer evaluation enables us to rigorously assess the robustness, portability, and domain-transfer capability of the learned representations across distinct clinical environments and patient populations. As shown in Figure~\ref{fig:mimic}, the transferred representations lead to stable and consistent AF prediction performance across logistic regression, random forest, and gradient boosting models. Notably, both supervised and unsupervised pre-training continue to outperform training from scratch in most cases, indicating that the hypergraph-based pre-training framework captures clinically meaningful structure that remains useful even when deployed on a separate EHR system with different coding practices and patient characteristics. An exception is the gradient boosting model trained from scratch, which performs competitively in this external setting. We hypothesize that this is because gradient boosting is more sensitive to fine-grained feature interactions and subtle patterns in the raw clinical feature space—patterns that may be partially smoothed or abstracted away in low-dimensional pre-trained embeddings. In addition, the MIMIC-IV dataset contains a substantially larger number of patients than our ESUS-AF cohort, reducing the data-scarcity challenges that typically hinder traditional machine learning models and enabling gradient boosting to more fully leverage the raw feature distribution. Overall, these results highlight the strong potential of our pre-training approach for learning generalizable and transferable EHR embeddings that can support multi-institutional clinical AI applications.

\section*{Conclusion}

In this work, we address the critical challenge of sample scarcity and high dimensionality in AF prediction among ESUS patients, which often limits performance in rare-disease contexts. To the best of our knowledge, we are the first to apply hypergraph pre-training for cross-cohort, cross-task transfer (from PSCI to AF prediction) in ESUS. Leveraging hypergraph-based pre-training, our framework captures complex patient--disease encounters and improves predictive accuracy, even with limited labeled data. This work demonstrates the value of pre-training and transfer learning in healthcare, offering a scalable strategy when large, annotated datasets are difficult to obtain. Beyond AF prediction, our findings highlight the potential of hypergraph representations to model intricate medical histories, supporting robust and generalizable machine learning in clinical practice.

\section*{Acknowledgements}
This research was partially supported by internal funds and GPU resources provided by Emory University, the U.S. National Science Foundation (Awards 2442172, 2312502, and 2319449), and the U.S. National Institutes of Health (Awards K25DK135913, RF1NS139325, R01DK143456, U18DP006922, and R01HL166233). 

% \section*{Ethical Considerations}

% References as numbers
\makeatletter
\renewcommand{\@biblabel}[1]{\hfill #1.}
\makeatother

% unstr is used to keep citation order
\bibliographystyle{vancouver}
\bibliography{main}  

\end{document}